\documentclass[runningheads]{llncs}
\usepackage[T1]{fontenc}
\usepackage{graphicx}
\usepackage{subcaption} 

\begin{document}
\title{Assessing the Emergent Symbolic Reasoning Abilities of Llama Large Language Models}
\titlerunning{Assessing the Emergent Symbolic Reasoning Abilities of Llama LLMs}
%
\author{Flavio Petruzzellis\inst{1} \and
Alberto Testolin\inst{1,2} \and
Alessandro Sperduti\inst{1}}
\authorrunning{F. Petruzzellis et al.}
%
\institute{Department of Mathematics, University of Padova, Padova, Italy \and
Department of General Psychology, University of Padova, Padova, Italy}
\maketitle              
\begin{abstract}
Large Language Models (LLMs) achieve impressive performance in a wide range of tasks, even if they are often trained with the only objective of chatting fluently with users. Among other skills, LLMs show emergent abilities in mathematical reasoning benchmarks, which can be elicited with appropriate prompting methods. In this work, we systematically investigate the capabilities and limitations of popular open-source LLMs on different symbolic reasoning tasks. We evaluate three models of the Llama 2 family on two datasets that require solving mathematical formulas of varying degrees of difficulty. We test a generalist LLM (Llama 2 Chat) as well as two fine-tuned versions of Llama 2 (MAmmoTH and MetaMath) specifically designed to tackle mathematical problems. We observe that both increasing the scale of the model and fine-tuning it on relevant tasks lead to significant performance gains. Furthermore, using fine-grained evaluation measures, we find that such performance gains are mostly observed with mathematical formulas of low complexity, which nevertheless often remain challenging even for the largest fine-tuned models.

\keywords{LLMs \and open source \and mathematical reasoning \and formulas \and ListOps \and arithmetic}
\end{abstract}

\section{Introduction}
Large language models (LLMs) featuring billions of parameters can exhibit sophisticated cognitive skills as a by-product of their design and training process rather than being explicitly trained to learn such skills. For example, after being trained on large unlabeled corpora of text with the objective of predicting the next word in a sentence, without additional fine-tuning, LLMs can be ``prompted'' to perform a remarkable variety of tasks, including two-digit arithmetic, question answering, text summarization, and language translation \cite{brown2020language}.
Mathematical reasoning is still considered a challenge for most LLMs \cite{testolin2024can}, but it can be partially elicited in large models using prompting techniques such as chain-of-thought prompting \cite{DBLP:conf/nips/Wei0SBIXCLZ22}. However, whether symbolic reasoning can be considered an emerging ability is still a matter of intense debate \cite{schaeffer2023emergent,wei2022emergent}.

In this work, we study the symbolic reasoning abilities of various large language models, focusing on open-source models from the Llama 2 family \cite{DBLP:journals/corr/abs-2307-09288}. 
We adopt a symbolic reasoning benchmark consisting of synthetic datasets of mathematical formulas characterized by the possibility of manipulating problem difficulty at a fine-grained level. Although such problems may appear, and indeed are, distant from typical applications of LLMs, they can be effectively used to systematically study the reasoning skills and the limitations of these models. 

We tested Llama 2 models of different sizes, as well as MetaMath \cite{DBLP:journals/corr/abs-2309-12284} and MAmmoTH \cite{DBLP:journals/corr/abs-2309-05653}, two versions of the base Llama 2 model fine-tuned to solve mathematical problems.
In our experiments, we observe that Llama 2 models become more capable of solving symbolic reasoning problems as their size grows and that fine-tuning on domain-specific problems can further improve their performance.
At the same time, by carefully analyzing how performance improvements are related to problem difficulty, we find that the emergence of symbolic reasoning is mainly observed when models (especially fine-tuned ones) are probed with relatively simple formulas.

\section{Related works}
The concept of emergence  \cite{anderson1972more} has been successfully used to characterize the dynamics of neural networks \cite{hopfield1982neural} and describe cognitive phenomena in terms of self-organizing principles \cite{mcclelland2010emergence,zorzi2018emergentist}. In recent years, it has attracted the interest of the AI community following the observation that unexpected cognitive abilities could emerge by simply increasing the size of deep learning models \cite{wei2022emergent}.

Indeed, transformer-based models from the GPT family were first shown to exhibit remarkable abilities to perform new tasks from textual instructions or from a few examples, while only being trained to autoregressively predict the next word in a sentence \cite{brown2020language}.
Since then, several works have shown that LLMs can carry out many new tasks that were not explicitly included in their training data, including solving arithmetic operations \cite{brown2020language} and performing commonsense reasoning \cite{wei2022emergent}. The ability of these models to reason about novel problems ``zero-shot'', that is, without any direct training, has been directly compared to the human ability to reason by analogy \cite{webb2023emergent}. 

Reasoning abilities are generally tested in LLMs using commonsense reasoning tasks, math word problems and symbolic reasoning benchmarks \cite{DBLP:conf/nips/KojimaGRMI22,DBLP:conf/iclr/0002WSLCNCZ23,DBLP:conf/nips/Wei0SBIXCLZ22}.
The latter is a class of synthetically generated problems whose difficulty can be systematically manipulated by deriving complex problems from the composition of simple ones.
One example is the ``last letter concatenation'' task \cite{DBLP:conf/nips/Wei0SBIXCLZ22}, in which the model is asked to output the string of characters obtained by concatenating the last character of the words in a list (which can be arbitrarily long). 
Another one is the ``coin flip'' problem \cite{DBLP:conf/nips/Wei0SBIXCLZ22}, a version of the parity problem in which the model should track the state of a coin after a variable number of flips.

A significant boost in performance on these benchmarks has been obtained using chain-of-thought prompting \cite{DBLP:conf/nips/Wei0SBIXCLZ22} and its variants \cite{DBLP:conf/nips/KojimaGRMI22,DBLP:conf/iclr/0002WSLCNCZ23}.
These prompting methods let the model produce a sequence of reasoning steps through which it can exploit the knowledge contained in the prompt to gradually derive the final answer.



\section{Methodology}


In this work, we focused on a class of challenging reasoning problems in which the model is tasked to solve symbolic mathematical formulas.
As in other symbolic reasoning benchmarks used to test Large Language Models, the problems we examined do not require verbal reasoning to be solved.
Instead, they require the ability to discover and systematically apply an algorithmic reasoning procedure that should enable generalization to problems of any level of difficulty.

We considered symbolic formulas that can be nested, which means that each operand in a formula can in turn be another formula. This implies that, in general, several reasoning steps might be required to solve a given formula.
Indeed, similarly to other symbolic reasoning tasks, any formula can be solved by the iterative application of a simple rewriting rule. For example, the arithmetic expression \mbox{\tt (12+(3-(4+5)))} can be solved by first identifying a solvable sub-expression, i.e. \mbox{\tt (4+5)}, and then substituting the result obtained by solving that sub-expression, namely {\tt 9}, into the original expression, obtaining a simpler expression, i.e. \mbox{\tt (12+(3-9))}, which can be further simplified by iteratively applying the same procedure.
We note that, given this problem structure, chain-of-thought prompting could in principle allow LLMs to solve these tasks by decomposing each formula and iteratively solving its simplest components.

Since the complexity of formulas can be characterized in terms of two parameters, that is, the nesting level and the number of operands involved, the reasoning abilities of LLMs can be analyzed in a finer-grained way compared to other reasoning benchmarks. This allowed us to compare the performance of models of different sizes on problems of varying levels of difficulty and thus investigate the dynamics of emergence of symbolic reasoning abilities.

\subsection{Symbolic formulas}
We considered two types of formulas: operations on lists of single-digit integers derived from the ListOps dataset \cite{DBLP:conf/naacl/NangiaB18} and arithmetic expressions \cite{petruzzellis2023hybrid}.

The ListOps dataset was introduced to evaluate the capacity of neural networks to build parse trees of nested formulas.
The original dataset included formulas composed of operations on lists of integers, including minimum, maximum, median, and sum modulo $10$ of a list of integers.
To reduce the complexity of the problem, we only used minimum, maximum, and sum modulo $10$.
We also built data splits of ListOps formulas whose level of difficulty could be precisely characterized.
In particular, we made it possible to specify the number of arguments that appear in formulas at any level, and we fixed the number of nesting points at each level of a formula to two, as in the case of arithmetic formulas (see below).
We evaluated models on ListOps formulas that had two to four operands and one to four nesting levels.
Furthermore, we slightly modified the original format of the ListOps formulas by using a more explicit functional notation since it has been observed that the notation used to represent symbolic formulas can strongly influence the performance of transformers on arithmetic tasks \cite{DBLP:journals/corr/abs-2102-13019}.
For example, the formula \texttt{[MAX 3 9 1]} was rewritten in the new format as \texttt{MAX(3, 9, 1)}, since this notation is more likely to be observed in other mathematical datasets used for training and fine-tuning of the LLMs considered here.

For the arithmetic task, we generated arithmetic expressions with sum, subtraction, and multiplication operations between two integers sampled in the interval $[-99, 99]$.
We considered formulas with one to four nesting levels.
For each nesting level, a formula was nested in two points: that is, exactly two operands on that level could be other formulas.
In this work, we were more interested in testing the capacity of Large Language Models to systematically execute a sequence of operations, rather than their mathematical competence in arithmetic with multi-digit numbers \cite{cognolato2022transformers}. Therefore, when computing the final values of formulas we used the modulo $100$ of the intermediate results, as done in previous work on systematic generalization in transformers \cite{DBLP:conf/iclr/CsordasIS22}.

\subsection{Models}

We evaluated the symbolic reasoning abilities of three models: Llama 2 Chat \cite{DBLP:journals/corr/abs-2307-09288}, and two fine-tuned versions, MAmmoTH \cite{DBLP:journals/corr/abs-2309-05653} and MetaMath \cite{DBLP:journals/corr/abs-2309-12284}.
For all of them, we considered three model sizes with 7B, 13B, and 70B parameters.

The Llama 2 Chat model is a large language model optimized for dialogue use cases, trained on a mix of publicly available data. It generally performs better than existing open-source models, approaching some of the most powerful closed-source models on a series of safety benchmarks, and it achieves high performance on a variety of tasks ranging from common sense reasoning to world knowledge, reading comprehension, and mathematical problem solving \cite{DBLP:journals/corr/abs-2307-09288}.

We also chose to test two recently proposed fine-tuned versions of Llama 2, MetaMath and MAmmoTH, that were designed to improve the mathematical reasoning abilities of the base model using different fine-tuning strategies.
MetaMath has been fine-tuned on MetaMathQA, a companion dataset created by bootstrapping samples in the GSM8K and MATH datasets by rephrasing both questions and answers with the aim of increasing variety in the training samples \cite{DBLP:journals/corr/abs-2309-12284}.
MAmmoTH was created with the aim of generalizing to many different mathematical and reasoning domains, hence it has been fine-tuned on eight different popular benchmarks and evaluated on both in-domain and out-of-domain problems from different datasets \cite{DBLP:journals/corr/abs-2309-05653}.

\subsection{Prompting Strategies}
\label{sec:met-prompt}
In order to elicit the emergence of reasoning abilities in Llama 2 Chat, we have initially tested zero-shot chain-of-thought prompting, a recently proposed prompting method that was shown to achieve similar performance as chain-of-thought prompting without the need to craft exemplars \cite{DBLP:conf/nips/KojimaGRMI22,petruzzellis-etal-2024-benchmarking-gpt}.
However, we observed that Llama 2 Chat already produced reasoning steps in the output using zero-shot prompting, presumably as the result of fine-tuning with reinforcement learning through human feedback \cite{DBLP:conf/nips/Ouyang0JAWMZASR22}, and that zero-shot chain-of-thought prompting did not further improve the model's performance.
Therefore, we opted for a simpler zero-shot prompting strategy, in which we briefly describe the task and then directly ask the model to solve it. In the case of the ListOps dataset, we also briefly describe the semantics of the operators that appear in the expression. For example, a Llama 2 prompt to solve a ListOps formula could be the following: \textit{\texttt{MIN}, \texttt{MAX} and \texttt{SM} are operators on lists of single-digit integers which have the semantics of minimum, maximum and sum modulo 10, respectively. Solve the following expression involving these operators: \texttt{MAX(3, 9, 1)}. Give the final answer stating `The final answer is: \texttt{<NUMBER>}'}.

In the case of MAmmoTH and MetaMath, we instead used the official prompting strategy used by the authors of the model during fine-tuning.
The prompt formats for the two models are similar: they both include an initial sentence introducing a generic task to the model
followed by task-dependent instructions and a request to produce a response.
In our case, we used the Llama 2 zero-shot prompt as a description of the task to be solved and we inserted it in the official prompt format.
The MetaMath prompt contained an explicit request to solve the problem step-by-step, while this was not required with MAmmoTH models since they have been fine-tuned with reasoning problems solved via chain-of-thought prompting, and thus produce a sequence of reasoning steps by default.
For example, a MetaMath prompt to solve an arithmetic formula could be the following: \textit{Below is an instruction that describes a task. Write a response that appropriately completes the request. Instruction: Solve the following arithmetic expression: \texttt{((86+51)+(-74-35))}. Take the modulo 100 of intermediate values, i.e. keep the last two digits of the number with the sign. Give the final answer stating `The final answer is: \texttt{<NUMBER>}'. Response: Let's think step by step.}

To extract the final answer from the model response we used regular expressions to match integers and take the last one appearing in the text. 
We measured the performance of all models using sequence accuracy, which means that an output was considered correct only if it exactly matched the target.

\section{Experiments and results}
In this section, we present the results of our evaluations of the three models.
We first focused on the impact of model size on the performance, and then study how performance is modulated by different levels of problem difficulty.
Finally, we take a closer look at the errors committed by the models on the simplest examples, to better characterize their capacities and limitations.

\subsection{Model size improves global accuracy}
\label{sec:res-1}
\begin{figure}[t]
    \centering
    \includegraphics[width=\linewidth]{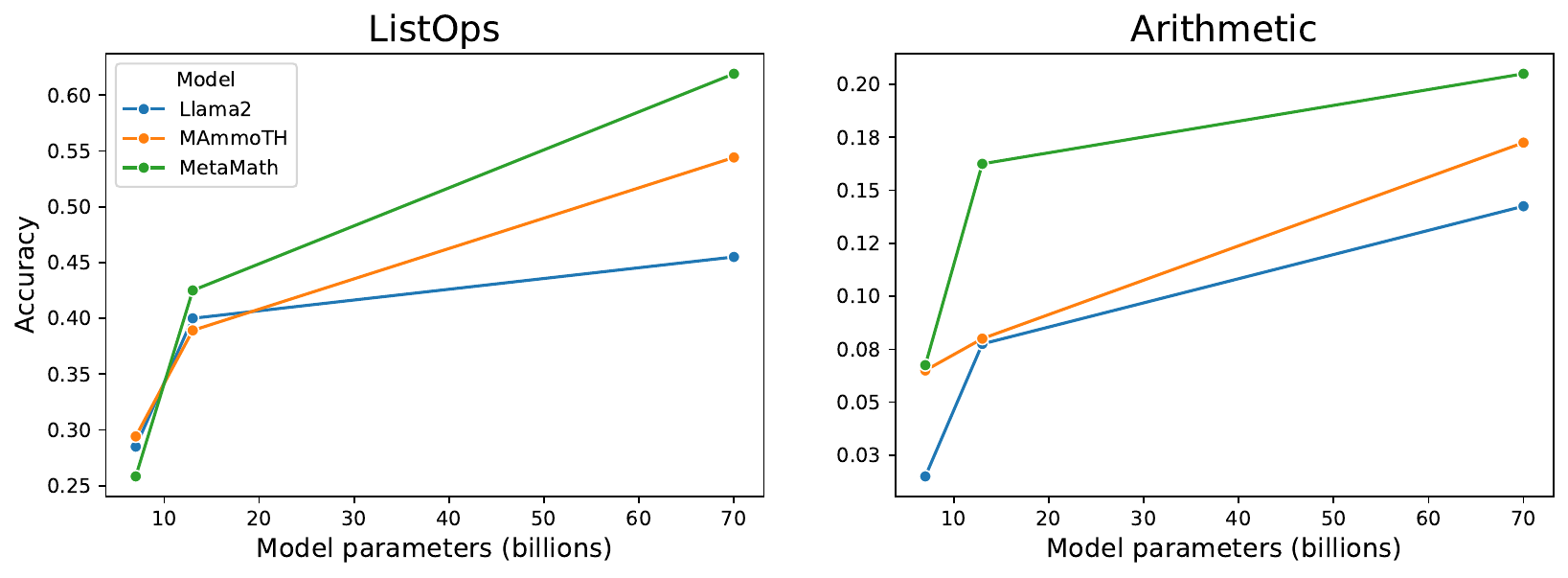}
    \caption{Average accuracy on the ListOps and Arithmetic tasks obtained by Llama 2, MAmmoTH, and MetaMath models of increasing size. Larger models (especially if fine-tuned on math tasks) achieve better performance than smaller ones.}
    \label{fig:aggr-acc}
\end{figure}

Fig.~\ref{fig:aggr-acc} shows the performance of the three models considered on ListOps and Arithmetic formulas, averaged across formulas of any nesting level and with any number of operands.
From this coarse-grained analysis, we can already observe that accuracy always increases with model size.
In particular, models with 13B parameters seem to develop significantly better symbolic reasoning abilities than their smaller 7B versions, and accuracy improves further in the largest models.

From the aggregate performance metrics, it is also clear that solving ListOps formulas is much easier than solving arithmetic ones, even for models that have been specifically tuned on mathematical reasoning datasets (note the different scale of the $y$-axes).
This phenomenon could be due to the fact that arithmetic formulas involve complex operations like multiplication between two-digit integers, and calculation of the modulo 100 of intermediate results.

It is also evident that fine-tuning Llama2 on mathematical problems improves its capacity to solve the kind of math-based symbolic reasoning problems used in our experiments. MetaMath models (especially the medium and large versions) outperform MAmmoTH models on both tasks, probably because they have been fine-tuned only on mathematical problems that are more similar to the ones we consider here.
Furthermore, in the case of ListOps, we find that fine-tuning is especially effective in boosting the performance of 70B models, leading to gains of 9\% (MAmmoTH) and 17\% (MetaMath) compared to the 70B Llama 2 model.
This could indicate that in the case of tasks consisting of a composition of relatively simple elementary operations, greater performance improvements can be achieved by fine-tuning large models.


\begin{figure}[t]   
    \includegraphics[width=\linewidth]{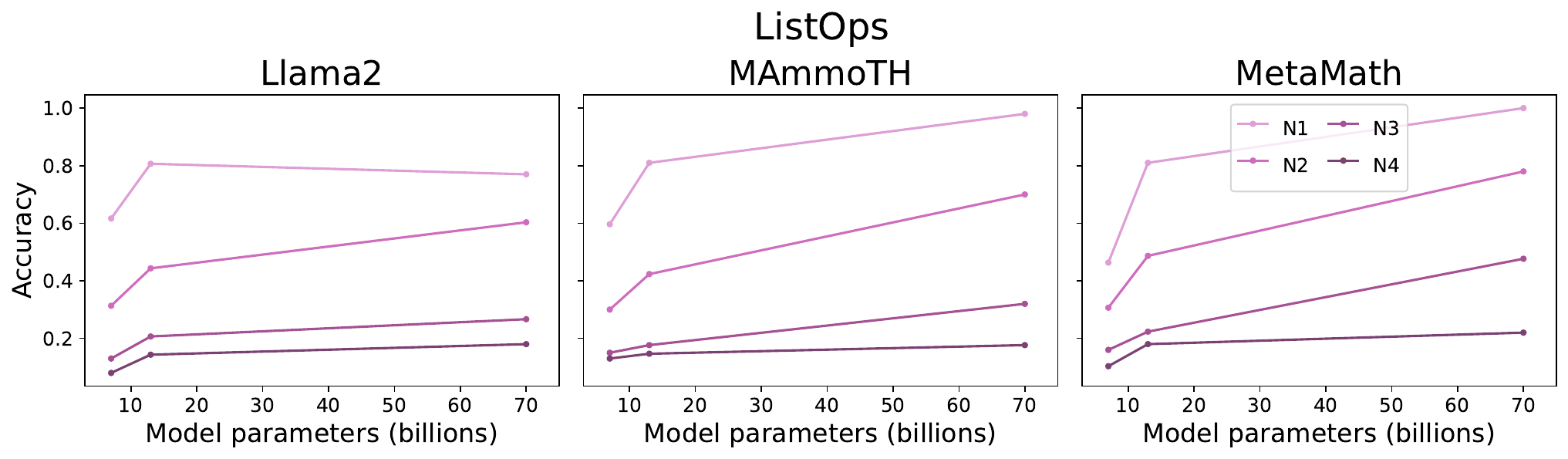}
    \includegraphics[width=\linewidth]{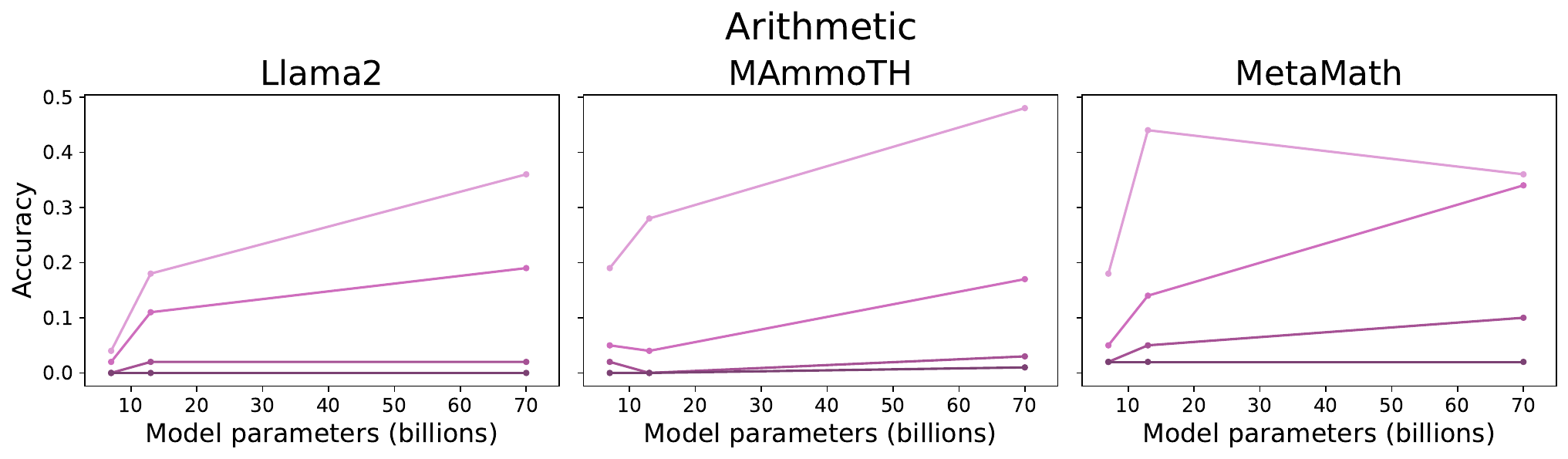}\caption{Accuracy of Llama 2, MAmmoTH and MetaMath models on ListOps (top) and arithmetic (bottom) formulas of varying levels of difficulty as a function of model size. \texttt{N$k$} indicates formulas with nesting level $k$.}
    \label{fig:acc-finegrain}
\end{figure}

\subsection{Accuracy improves more on simple formulas}
We now aim to better understand where the performance gain obtained by larger models is concentrated. 
In Fig.~\ref{fig:acc-finegrain}, we report a fine-grained measure of the accuracy of the three models on groups of ListOps and arithmetic formulas of increasing nesting levels.
As expected, all models are generally more accurate when solving formulas in the easiest splits of both tasks (i.e., formulas with only one or two nested expressions).
This indicates that increasing the nesting level of a formula indeed makes the problem more difficult for all the models.

We further notice that as model size grows, accuracy generally increases more on formulas with nesting levels 1 and 2, both for Arithmetic and ListOps.
For ListOps we observe a slight improvement in accuracy with model size also for the most complicated formulas, while for the most challenging arithmetic problems the improvement is almost null.
It is also interesting to notice that the largest version of MetaMath (70B) is slightly less accurate than the intermediate version (13B) in arithmetic problems with a single nesting, suggesting that scaling-up model size might be detrimental in some cases \cite{DBLP:journals/corr/abs-2306-09479}.

These results suggest that the emergent symbolic reasoning abilities observed in the largest models do not yet allow for compositional generalization \cite{DBLP:journals/jair/HupkesDMB20}, being mostly effective on relatively simple formulas. This holds even for fine-tuned models, since their accuracy on the most challenging problems (four nested formulas) is comparable to that achieved by the base Llama 2 models.

\begin{figure}[t]
\begin{minipage}[c]{0.48\linewidth}
    \includegraphics[width=\linewidth]{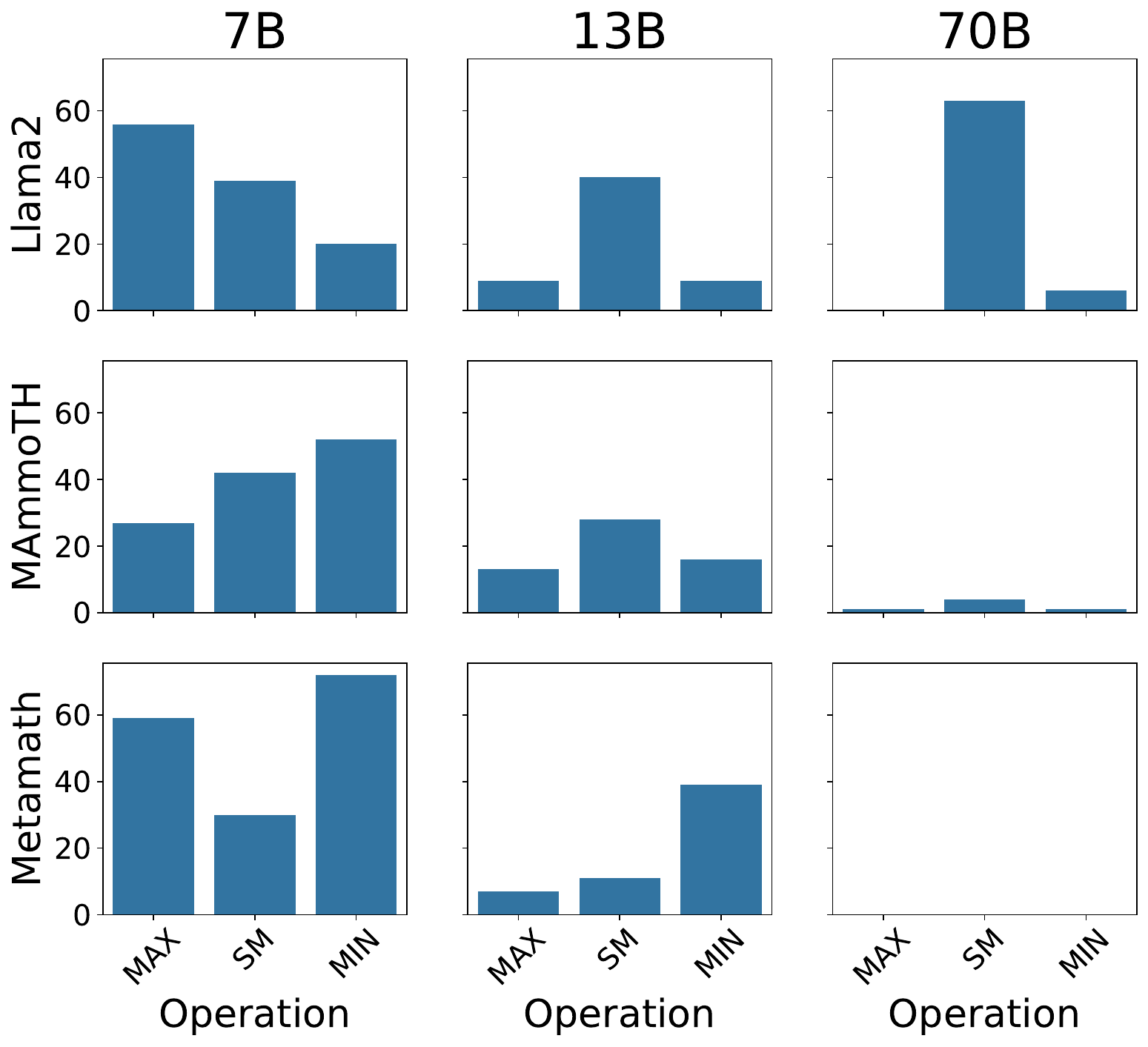}
    \caption{Type of errors made by the models on ListOps formulas with a single nesting level. Absolute number of errors is on the $y$-axes and operator used in the formula is on the $x$-axis.}
    \label{fig:error-by-op-list}
\end{minipage}
\hfill
\begin{minipage}[c]{0.48\linewidth}
    \includegraphics[width=\linewidth]{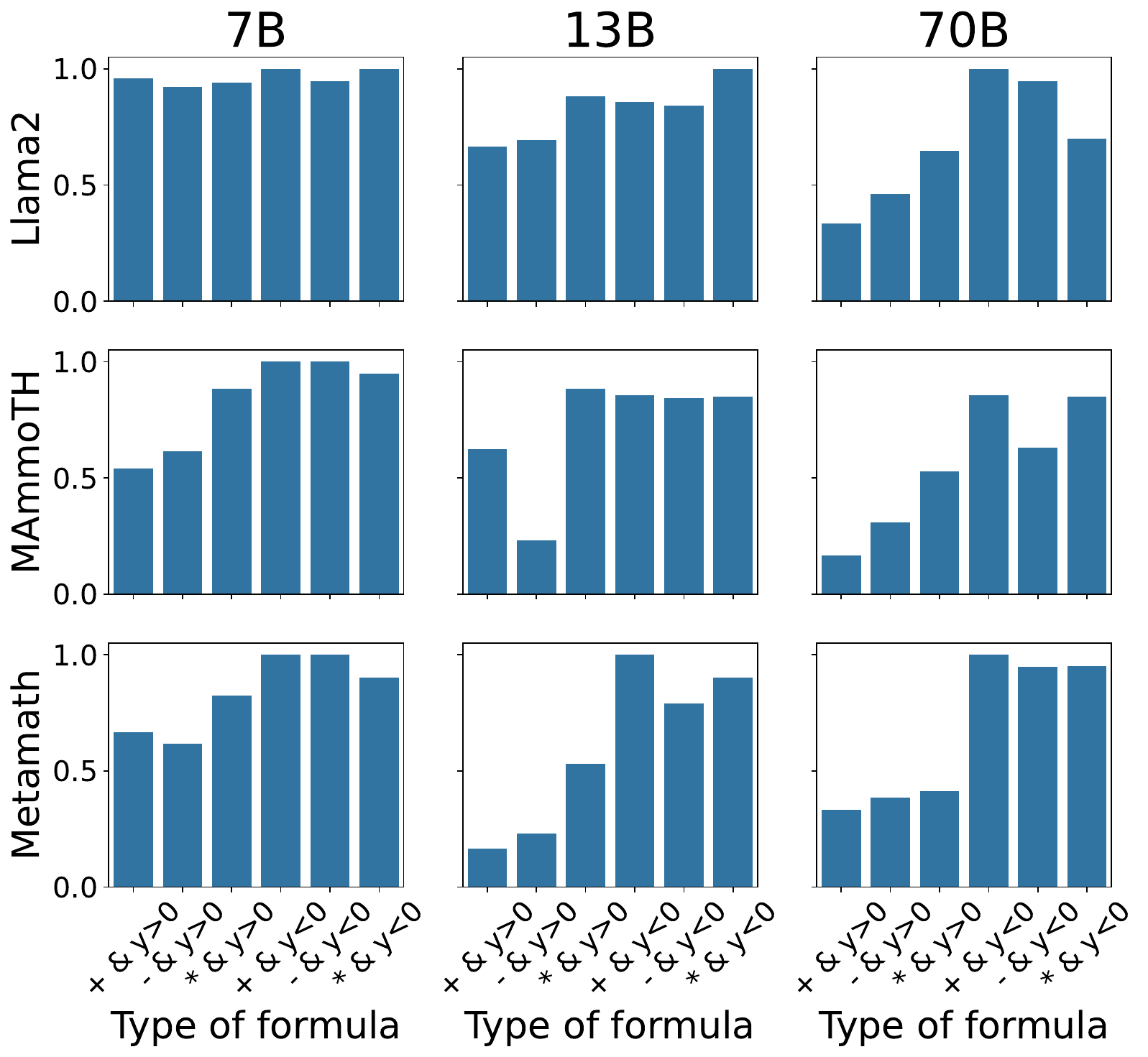}
    \caption{Type of errors made by the models on arithmetic formulas with nesting level 1, grouped by operator (+,-,$\ast$) and sign of the result ({\rm y}). Incidence of errors is measured by group.}
    \label{fig:error-type-arit}
\end{minipage}
\end{figure}
\subsection{Analysis of errors on simple formulas}

In order to solve the symbolic reasoning problems considered in our experiments, the models should be able to solve both atomic operations, such as summing two numbers or finding the maximum in a list, and apply the correct sequence of solution steps in nested formulas.
Since we generally observed that performance mostly improved on simpler problems, in the following analysis we focus on studying the capacity of the models to solve the simplest arithmetic and ListOps formulas.
To this aim, we isolated and analyzed the errors on data splits with a single nesting level.

In Fig.~\ref{fig:error-by-op-list}, we show the number of errors committed by the models on ListOps formulas with a single nesting.
We observe that for Llama 2, the scale of the model allows to significantly improve its ability to solve min and max operations (almost to perfection), while the sum modulo 10 becomes the most difficult operation in the largest model, with a surprising deterioration in performance compared to smaller-scale models.
The trend is different for fine-tuned models, which have presumably observed the sum modulo 10 operation more frequently during training and can thus solve it better than the base model, with the largest model versions reaching almost perfect accuracy on all atomic operations.

In Fig.~\ref{fig:error-type-arit}, we show the mistakes made by the models on arithmetic formulas with a single nesting level.
We group input samples based on the type of operation appearing in the formulas and on the sign of the result, as we hypothesize that operations involving negative operands could be more difficult than those on positive ones.
We then measure the incidence of errors in each group, i.e. the fraction of formulas in each group that the model does not solve correctly.
We observe that both fine-tuning and reasoning abilities emerging with scale mainly improve the models' accuracy on formulas that have a positive result.
By looking at the reasoning steps produced by the models, we noticed that the vast majority of these errors are due to an incorrect calculation of the modulo operation, which indeed proves to be more difficult for all models when it involves negative operands.

\section{Conclusions}
Despite the widespread deployment of foundation models, we currently lack a clear understanding of how they work, when they fail, and what are the capabilities and limitations of their seemingly emergent cognitive skills \cite{schaeffer2023emergent}.

In this work, we studied the emergent symbolic reasoning abilities of open-source LLMs of the Llama family on mathematical formulas whose level of difficulty can be precisely manipulated. 
We considered Llama 2 Chat and two variants of the model fine-tuned for mathematical reasoning, comparing the performance of small, medium, and large versions of each model.

We found that larger models are generally more capable of solving mathematical formulas compared to smaller ones, suggesting the emergence of symbolic reasoning abilities.
However, a finer-grained analysis revealed that accuracy mostly increased on formulas of low complexity, involving only a few nesting levels.
Furthermore, by analyzing the models' failures on such simple formulas, we found that common expressions like the modulo operation can still represent a challenge even for the largest and fine-tuned language models.

Overall, our results suggest that large language models still struggle in tasks requiring symbolic reasoning, and further research is needed to design neural architectures better suited for this type of tasks.
While our findings are limited to models in the Llama family, we believe that the proposed evaluation approach, based on symbolic reasoning benchmarks in which the difficulty of samples can be precisely characterized, can help to build a more sound understanding of the potential and limitations of this technology.

%

\bibliographystyle{splncs04}
\bibliography{bibliography}

\end{document}